\documentclass[journal]{IEEEtran}
\usepackage{ifpdf}
\usepackage{setspace}
 \usepackage[final]{changes}
\usepackage{graphicx}
\usepackage[space]{grffile}
\usepackage{latexsym}
\usepackage{textcomp}
\usepackage{comment}
\usepackage{longtable}
\usepackage{tabulary}
\usepackage{booktabs,array,multirow}
\usepackage{amsfonts,amsmath,amssymb}
\usepackage{url}
\usepackage{hyperref}
\hypersetup{colorlinks=false,pdfborder={0 0 0}}
\usepackage{etoolbox}
\usepackage{subfigure}

\newif\iflatexml\latexmlfalse

\AtBeginDocument{\DeclareGraphicsExtensions{.pdf,.PDF,.eps,.EPS,.png,.PNG,.tif,.TIF,.jpg,.JPG,.jpeg,.JPEG}}

\usepackage[utf8]{inputenc}
\usepackage{cite}
\usepackage{balance}

\begin{document}


\title{The \textit{Why}, \textit{What} and \textit{How} of Artificial General Intelligence Chip Development}


\author{Alex James, \it{IEEE Senior Member}% <-this % stops a space
\thanks{A. P. James is with Kerala University of Digital Sciences, Innovation and Technology (KUDSIT), and Maker Village, Indian Institute of Information Technology and Management, Kerala. Email: apj@ieee.org}% <-this % stops a space

}% <-this % stops a space


\markboth{}%
{}



\maketitle


\begin{abstract}

The AI chips increasingly focus on implementing neural computing at low power and cost. The intelligent sensing, automation, and edge computing applications have been the market drivers for AI chips. Increasingly, the generalisation, performance, robustness, and scalability of the AI chip solutions are compared with human-like intelligence abilities.  Such a requirement to transit from application-specific to general intelligence AI chip must consider several factors. This paper provides an overview of this cross-disciplinary field of study, elaborating on the generalisation of intelligence as understood in building artificial general intelligence (AGI) systems. This work presents a listing of emerging AI chip technologies, classification of edge AI implementations, and the funnel design flow for AGI chip development. Finally, the design consideration required for building an AGI chip is listed along with the methods for testing and validating it. 
\end{abstract}%


%\begin{IEEEkeywords}
%IEEEtran, journal, \LaTeX, paper, template.
%\end{IEEEkeywords}

% \fi
\IEEEpeerreviewmaketitle

% *** Do not adjust lengths that control margins, column widths, etc. ***
% *** Do not use packages that alter fonts (such as pslatex).         ***
% There should be no need to do such things with IEEEtran.cls V1.6 and later.
% (Unless specifically asked to do so by the journal or conference you plan
% to submit to, of course. )

\section{Introduction}

{\label{880455}}

%\subsection{Motivation and Need}
\IEEEPARstart{T}{he} 
 quest for building an artificial brain developed the field of computers and artificial intelligence. The machine that could mimic human intelligence has been proven to be a hard problem. Most of the general intelligence problems are AI-hard~\cite{Yampolskiy_2013}~and require
the rethinking of AI architectures and hardware if artificial general intelligence (AGI) chip were to be developed.

The weak-AI~\cite{Kim_2017a} is defined as those AI systems that are designed for specific applications. For example, the task of face recognition and detection is a weak-AI task. The same AI system cannot be used for another application without retraining and at a time can
only process a handful of tasks. Most of the current AI systems attempt weak AI problems. There is a lot of practical use and application of the weak-AI systems, and they are the first step towards realizing general
intelligence.

%The weak-AI systems do not have self-awareness~\cite{Chella_2020,Lewis_2014}. While many weak-AI systems show excellent performance based on specific rules and tasks, they do not perform well when it comes to real-time situations unforeseen to the system.~ For example, the weak-AI systems, apart from being limited in capabilities when used in autonomous
%vehicles, the system while trying to achieve performance do not consider the mindset of the human who might be purposefully trying to harm the vehicle through collision, thereby ending up miscalculating the
%directions.

In contrast, the strong AI systems~\cite{Voss,Pennachin,Yudkowsky} aim to overcome the limitations of weak-AI by incorporating general intellectual
abilities. The environmental awareness and innate knowledge, along with learning abilities enable understanding of various rules, policies, social behavior to ethical aspects. This target abilities make the best benchmark known today for
strong AI systems to be that of human intelligence.

%The large computations that are required to make collective decisions based on events and tasks are usually envisioned to be implemented in a high-performance computer. While sensory detection and processing are done within the machine itself. The ability of the machine to remain intelligent requires good communication bandwidth and requires efficient ways to manage power.

The competition to build an AGI system in a machine is largely
approached in algorithmic and software ways using traditional digital
computers and servers~\cite{Goertzel_2009,Pennachin,Voss,Red_ko}. {The recent approaches such as using Z$^*$-numbers \cite{8845755}  provides an architectural and algorithmic framework for  realising real-time perception, and attention functions. Such approaches are relatively easier to translate to digital systems, while the analog/mixed-signal implementations could require extensive hardware design investigations.  } Building autonomous AGI machines even with existing low power AI chips would require complex interface electronics, high computing capacity, and infrastructure. Suppose, we would like to build a robot as a football player that has AGI capability. The primary concerns would be that of energy efficiency, real-time response times, fast decision making skills, and agility. Increasingly, many such functions are achievable with biomimicry and sensing control circuits, however, they do not match up with equivalent energy efficiency, robustness, generalisation ability and adaptive behaviour of biological systems. 

{The energy-efficient AI chips build by mimicking neural networks and with many emerging devices have shown promising results\cite{Camu_as_Mesa_2019,Zyarah_2018,Gale_2019,Itoh_2019, pei2019towards}. Most, if not all, AI chips on the market today including those aim towards early AGI natives are good at solving one or other weak-AI problems, at much lower energy and on-chip area compared to server-based alternatives. Since the processing is done within the edge devices, the response times improve and it becomes closer to being suitable for real-time use. However, they do not address the issue of general intelligence in hardware, and often address this question today from a philosophical and algorithmic perspective. {In this process, the AI chip is often seen as having an accelerator or speed role \cite{chen2020survey}, rather than seeing it as the primary building block for general intelligence.}}

{This paper presents a critical overview of the emerging field of
hardware AI focusing on developing AGI
chips. The origin of general intelligence is presented relative to the understanding of intelligence from a cross-disciplinary perspective. The concept of edge intelligence, followed by the development
approaches for AGI chip and how it can be tested, is proposed. This work's primary goal is to bring in a convergence of important ideas from AI research that can be useful for AGI chip development. We ask the questions of \textit{Why} and \textit{What} to connect the theory of intelligence with that of AI hardware, placing this as building principle blocks of AGI chips. The question of \textit{How} aims to uncover the practical aspect of designing, testing, and validating the AGI chips when developed.}

{The
paper is organized into nine sections. 
Section~{\ref{368047}} introduces the concept of intelligence and its linkages to building AI machines.~
Section~{\ref{352365}} gives an overview of edge AI
differentiating and categorizing it based on functional executions of AI
operations. Section~{\ref{738216}} and V provides insight
into building blocks for an AGI system.~
Section~{\ref{321616}} introduces a law of funnel
design in AGI chip. Section~{\ref{960251}} proposes how
AGI chips can be tested, Section~VIII provides discussion and classification of AGI development. In summary, Sections II-III addresses the question of \textit{Why} and \textit{What}, while the Section IV-VIII addresses the question of \textit{How} in AGI chip development. Finally, Section~{\ref{900499}}
provides the main conclusions. }

\begin{comment}

\section{Literature Review}

\paragraph*{Why this survey?}

The interest in the development of AI chips are on growth. The main purpose if this review is to bring together various concepts for AGI with context to existing literature of AI theory and hardware. Broadly, the area of AI has progressed from the idea of replicating neural models to high density neural chips capable of speeding up neural computations and its applications. The ability to replicate higher level of cognition that makes the human intelligence superior that machine intelligence is one of the main goals of AGI. Given this is a natural progress to the next stage of AI systems, this survey places the views on how such AGI chips could be developed and how it can be tested once realised. 

\paragraph*{Who is this survey intended for?}

\subsection{Methodology}

include a Survey Methodology section (see https://peerj.com/about/author-instructions/#literature-review-sections).

You should use this section to explain how you ensured comprehensive and unbiased coverage of the literature. This must include search engines and search terms used, criteria for inclusion/exclusion of articles, etc.

\end{comment}

\section{Towards Intelligence in Hardware}

{\label{368047}}

{This section aims to provide a broader context and meaning of intelligence inspired by human brain understanding (Section \ref{368047}\textit{A-B}). Further, the transition of intelligence understanding useful for AGI  is reflected through the recent efforts on implementing AI chips and emerging hardware systems (Section \ref{368047}\textit{C-D}). }

\subsection{Definition of Intelligence}

{Different definitions of intelligence exists today in the literature. There are variations in the way intelligence is perceived by one scientific community to another, with no formal  definition that is universally accepted. In context of machine intelligence, Legg and Hutter's \cite{legg2007collection} definition -  "\textit{Intelligence measures an agent's ability to achieve goals in a wide range of environments.}" - stand-out as a prominent one that in essence defines intelligence as an agent's ability to perform a set of task and achieve goals in a diverse set of environments.}

The agent can be implemented in the form of a software or hardware or more likely a combination of both in AI hardware perspective. Such agents if they were to survive diverse challenges would need the ability to reason, plan, solve problems, think abstractly, comprehend complex ideas, learn quickly and learn from experience. {In essence these agents will require the ability to sense, react and respond to the environmental conditions. This would mean, the agent should be able to generalise, understand and act upon a wide range of situations, typical of intelligent decision making capability of humans.} This form of general intelligence originates from the ability to comprehend and survive in diverse set of environmental conditions. Further, individual tasks can be executed in logic operations and rules programmable with minimalist learning. In cognitive studies, the idea of generality and purpose drive intelligence map to the theory of fluid and crystallized intelligence {(as explained in chapter five of the book \cite{cattell1987intelligence})}, while, holding a deeper link and connection to the ideas of mind.  There are two floating ideas of mind, (1) being bounded by evolution that makes it very deterministic and with limited ability to learn, and (2) being an unbounded general purpose system that can learn and solve any problem (see pp. 89 in \cite{spelke2007core}).

{The agents ability to perform a set of tasks} are relatively easier to quantify and evaluate in AI studies, that can be implemented as application driven assessments. {On the other hand, the idea of} generality however, is unbounded by the conditions presented to the intelligent system, and are harder to quantify and evaluate.

\subsection{Understanding Intelligence}
{There are in totality three different yet broad approaches that have been followed in the past to understand intelligence. They are, (1) based on the principle of evolution, (2) that of philosophy of generality, and (3) as a result of studies in developmental and cognitive psychology.}

\subsubsection{Evolution-driven intelligence} {The idea that intelligence is evolution driven (as explained in  \cite{mcnally2012cooperation, dunbar1989machiavellian}), and general theory on that ability to solve a problem is derived from the  task specific adaptions \cite{kanazawa2004general} makes it a compelling case for machine intelligence. As pointed out in \cite{kanazawa2004general} (pp. 514) - "\textit{then any genetic mutation that equips its carrier to think and reason would be selected for and could evolve as a domain-specific adaptation in order to solve novel, nonrecurrent problems}", this, in essence, one can think this as divide and conquer approach where a complex task is divided into several sub-tasks, allowing domain specific evolution. If an AI system can solve each sub-task individually, it is assumed that it can solve the cumulatively resulting complex task.}

In terms of AI hardware, this can be considered as a modular approach to build higher form of intelligence. For example, each AI circuit block within a chip could be fine-tuned to solve a specific problem such as image recognition, object recognition, and speech recognition. Several such blocks together forms an AI chip that can then collectively solve a complex problem that make use of multi-modal analysis and decisions performed by individual blocks. The task specific adaptations in this case would need memory and rules to make it work, which should be provided by humans. The idea of learning in this case is highly simplified to specific tasks under strict constraints. { This idea is in firm alignment with that of the Minsky's view of AI (for example chapter 3-4 in \cite{minsky1972artificial}), in explaining language processing and semantic understanding. Where the science of building AI was coupled to the set of tasks the humans program it to do.} Hence, indirectly mapping the human skills and knowledge to a deterministic and organised representations of rules and memory becomes the norm.

\subsubsection{Generalisation-driven intelligence}

{The idea of generalisation gives a bigger importance to adaptation and learning. In this approach, the intelligence development is envisioned as the ability of the system to learn new tasks, and solve problems that are unknown previously to the system. The pioneers like Turing, McCarthy and Papert were first few to extensively envision the generalisation view of intelligence \cite{mccarthy1987generality, minsky1972artificial}.} {This is pointed out by McCarthy during the 1971 Turing Award Lecture \cite{mccarthy1987generality} "\textit{In my opinion, getting a language for expressing general commonsense knowledge for inclusion in a general database is the key problem of generality in AI}".}

{Until the development of machine learning and deep learning systems, the evolutionary-driven intelligence dominated the field of AI. In present times, there is hardly any disagreement on the need to approach AI through learning and assess its ability to adapt to unknown problems. However, the immediate requirements in the AI industry are task specific applications, that often do not need high generalisation ability.}

{The inculcation of many ideas from the past support towards this. The foremost from the AI literature is the ancient idea of Tabula Rasa \cite{lawson1988acquisition}, where the mind is considered as a blank slate, and anything can be written on it that is enriched by experiences in time. An untrained neural network can be considered as a 'Tabula Rasa' in the sense that it can be trained to perform a given task through weight updating algorithms. Although, it may make sense to establish this correlation, it is an incorrect assumption given the various progress in the developmental studies related to intelligence contrast this.}

\subsubsection{Human intelligence}

{Several studies from developmental psychology point out that both  evolution and blank slate only approach to intelligence is wrong. Instead, the mind is capable of high degree of generality, not limited by innate skills and capable to acquire new skills throughout the life. At the same time, our cognitive abilities are specialised by evolution that is driven by biological priors and evolutionary limitations that dismantle 'blank slate' theories as well. }

{The fact that humans are capable of excelling in specific tasks and skills compared to many other animals are a testament to this. Understanding human cognitive priors are essential to developing human like intelligent forms. }

{There are several priors that help humans perform intelligent tasks \cite{versace2018priors}. They can be grouped as}:
\begin{enumerate}
    \item {Motor-sensory priors: These are motor-sensory skills that humans are born with such as ability to move, react and respond to sensory inputs. The reflexes such as vestibulo-ocular reflex is a good example of this, where the head movements coordinate strongly with the eye movements.}
    \item {Meta-learning priors: These are the underlying strategies what define how we learn any tasks. For example, the idea of functional modularity, hierarchical information processing, spatio-temporal information coding, decoding and organisation can be classified under this.}
    \item {Knowledge priors: These priors gives a broader sense of the environment around us. For example, the shapes of objects, innate indication about space and depth, innateness of numbers, sense of time, and sense of social intuitions.}
\end{enumerate}

{The development of AI hardware that cater to only motor-sensory priors such as using sensors and responses helps to imitate an artificial human body without general intelligence abilities. While, meta-learning priors form the key aspects of intelligence that enable learning and help solve problems, AI chips today make use of various forms of meta-learning priors, and are in constant effort to identifying newer and efficient learning implementations. In contrast, the knowledge priors help assess the AI chip implementations against human like intelligent tasks. The AI chips that could potentially include programmed number of knowledge priors could trick the specific test to show human like intelligence ability, however, they do not represent general intelligence, and such system should not considered as AGI systems.}

%\section*{\sout{Mind and Machine}}
%\subsection*{\sout{Human Mind}}

{\label{117500}}

\subsection{From General Intelligence to
AGI}

{\label{503281}}

%An individual is expected to demonstrate general intelligence when there is evidence of the ability to integrate multiple cognitive functions in a fluid manner~\cite{Voss,Bringsjord_2009}. A high level of general intelligence allows for solving new problems developing newer knowledge and higher reasoning abilities. Hence, the possibilities of general intelligence are much ahead and superior to weak AI systems of today~\cite{Pennachin,Bates_2002,Yudkowsky}.

{AGI aims to embed general intelligence
abilities in machines and are also known as 'full AI' or 'strong AI'
systems\cite{Pennachin,Bates_2002,Yudkowsky}. {There is no evidence of practical \emph{strong AI} systems that
exist today that matches fully with human like intelligence, although there have been active attempts to build
libraries and software (e.g. see \cite{goertzel2014opencog}) to build such systems.} While, these software and libraries are essential to progress the AGI, they alone cannot achieve physical compactness and energy efficiency required for such complex systems. This is where, AGI chip development becomes important, where either they become supportive tool as an accelerator to algorithms or that can have innate priors embedded within its hardware.}

{The knowledge priors are extensively used in testing AGI algorithms, by giving and comparing human-like challenges to AGI system. Examples include: (1) testing the AGI system by making it
take a university program and obtaining a degree, (2) testing the ability
of the AGI to perform in a job, (3) testing the ability of AGI system to
make a coffee, and (4) perform Turing test~\cite{2019,Neufeld_2020,Searle_2009,2019a,Yudkowsky}.}

The AGI system should be able to solve 'AI-hard'
problems~\cite{Andreotta_2020,1995}, that would require the use of natural
language processing, computer vision, and understand images,
emotions, and environment. With the
current state of the art AI system, solving an AI-hard problem requires
the assistance of humans in addition to the acceleration provided by
computing machines.

%There is growing speculation on whether AGI systems can be realized and not just remain as a concept. The speculations arrive from the increasing complexity and computational resources required for solving weak AI programs, let alone a progress towards strong AI system.

{Majority of AI methods aim to mimic the functional and/or
structural behavior of the neural networks. The functional understanding of brain is emulated through knowledge representation, statistical models and complex networks. Mimicking biology driven human intelligence will require the application of the concepts of feedback and feedforward propagation of information, and integration of sensing mechanisms, along with meta-learning priors and knowledge functions specific to tasks. }

%The concept of feedback and feedforward propagation of information, learning and updating the connection between the neurons, and layer by layer processing of information inspire the designs. Collectively, they are known as neuromorphic computing and neuron-inspired techniques such as reflected through deep learning architectures.

The brain is considered as the benchmark system for AGI research. There
are many approaches in neuromorphic computation that aim for achieving energy efficiency and performance benchmarks of the
brain~\cite{Zhu_2020,Roy_2019}. The performance of the AI systems has
considerably improved due to the availability of
high-performance computers and excessive capturing of labeled data.
As more and more sensors are connected to the devices, the volume,
veracity, and velocity of data increase. The deep learning architectures
\cite{Sengupta_2020} can use the data to arrive at improved
classification and recognition accuracy.

\begin{figure*}
\centering

\subfigure[]{\includegraphics[width=3cm]{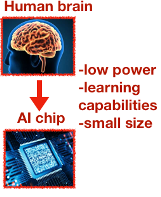}}
\subfigure[]{\includegraphics[width=15cm]{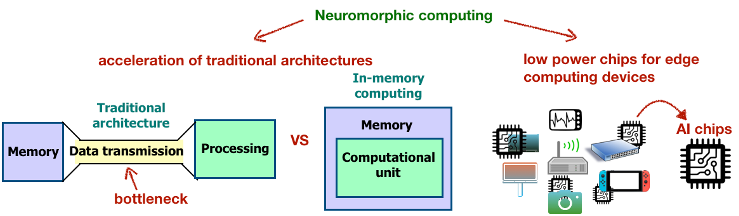}}
\subfigure[]{\includegraphics[width=6cm]{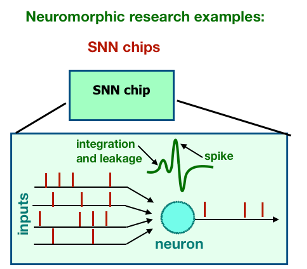}}
\subfigure[]{\includegraphics[width=10cm]{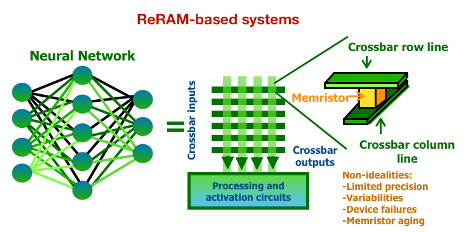}}

\caption{{An overview of hardware on-chip neural networks. (a) shows the key
benchmarks for AI chip, (b) illustration of in-memory computing and
applications, (c) SNN concept, (d) crossbar with ReRAM for implementing a layer of neural network.}
\label{654862}}
\end{figure*}

\subsection{AI Hardware}

{\label{898853}}

The neuromorphic computing is inspired by the information processing in
a human brain by mimicking the operation principle and structure of
the neurons and synapses (Figure~{\ref{654862}} (a)).
Low power intelligent information processing is a key concept in
neuromorphic computing inspired by the human brain which performs
complex processing tasks consuming nearly 20 W of power. The first
task-specific neuromorphic systems have been developed for specific
applications, such as object recognition or prediction of specific
data. The second generation of the neuromorphic systems is moving
towards the tasks corresponding to human cognition, like adaptation and
interpretation of the unknown information, and a final goal of building
strong AI systems\cite{Voorhees_1999,Kwok_2019}.

\subsubsection{Neuromorphic systems}

{There are numerous practical implementations of neuromoprhic chips. Some examples are HICANN, HICANN-X\cite{aamir2018accelerated}, SyNAPSE\cite{cruz2013scalable}, SpiNNaker, SpiNNaker-2\cite{partzsch2017fixed}, True North\cite{cassidy2014real}, Neurogrid\cite{benjamin2014neurogrid}, IFAT\cite{park201465k}, ROLLS\cite{indiveri2015neuromorphic}, DYNAP-SEL\cite{qiao2016scaling}, Loihi \cite{davies2018loihi} and SBNN\cite{chen20184096}. Besides, CMOS devices have gained popularity in the recent years, and its practical realisation is also on the raise.}

{The development of neuromorphic computing and architectures
is facilitated by the scaling of the transistors and development of the
emerging technologies, such as ReRAMs, PCMs, and
STT-RAM, allowing to increase the density of the
devices on-chip and reduce the power consumption \cite{ielmini2019emerging}. Neuromorphic computing
involves the acceleration of traditional computing architectures and the
development of low power AI chips useful for edge devices and
near-sensor processing (Figure~{\ref{654862}} (b)). The
acceleration of traditional architectures includes various accelerators,
like spike-based processors \cite{Kousanakis_2017}, and in-memory computing
architectures \cite{Mohamed_2020}.}
%In-memory computing architectures are proposed as a solution for faster computing, compared to the traditional computing systems where processing is performed separately from the memory and data transmission speed between memory and processor is limited.~

Among the emerging devices for building AI hardware, the class of non-volatile memory
memristive devices stands out, as they have useful properties that mimic
the neuron's electrical behavior~\cite{Cai_2014}. When a stimulus
voltage pulses with a constant magnitude and time period are applied
across the memristors over a period of time, the current output drops,
indicating that there is an adaptive behavior. This correlates with
single-neuron behavior, as it resembles neurons being able to learn a
stimulus. Multiple memristor neurons can be arranged in a
crossbar~architecture~\cite{Vourkas_2019,Vourkas_2014} and the voltage difference
across the devices carefully tuned, to emulate a spiking neural
network~\cite{Querlioz_2011,Zheng_2018}. Several configurations of the crossbar
architectures are possible to emulate various known artificial neural
networks such as multi-layered perceptron, deep neural network,
convolutional neural network, long short-term memory, and generative
adversarial networks.

The non-idealities of CMOS-memristive/emerging devices based crossbar
systems include the scalability issues and leakage currents,
non-linearity during programming, variabilities in resistive levels,
limited number of stable resistive states, aging issues, drifting of
resistive levels, device failures, endurance issues and limited lifetime
of real memristive devices~\cite{Van_Pham_2019,Zhang_2019,Krestinskaya_2019}. These make real-time
learning with emerging devices to be a nontrivial task. This also
implies the need to have smarter ways to train, program, and implement
algorithms.
\subsubsection{Digital neural chips}
The digital neural chips are the most established AI chip implementations for practical use. {The synaptic nodes become purely digital when the 'multiply and accumulate' (MAC) operation is implemented using digital logic.}   Some of the implementations are Diannao\cite{chen2014diannao}, Dadiannao\cite{luo2016dadiannao}, Pudiannao\cite{liu2015pudiannao}, Shidiannao\cite{du2015shidiannao}, Eyeriss\cite{chen2016eyeriss}, EIE\cite{han2016eie}, Origami\cite{cavigelli2015origami}, Envision\cite{moons201714}, TPU\cite{jouppi2017datacenter}, Tesla, DPU, Q4MobilEye, Parker, S32V234, and Myriad 2\cite{moloney2014myriad}. A distinct from CPU and GPU chips that are controlled by software algorithms, the digital neural chips uses dedicated hardware units for accelerating neural computations.

{The hardware performance of digital neural chips is assessed based on MAC operations per second (i.e. two floating point operations), throughput and clock frequency. Only about 10\% of area in the chip is occupied by neurons, while the rest is occupied by memory units and control circuits.}

\subsubsection{Other emerging systems}

{The carbon nanotube (CNT) along with thin film transistor (TFT) has been used to build synapsis. It is also feasible to fabricate 3D architecture of CNT TFTs logic with vertical metal oxide memristors (as non volatile memory (NVM)) and silicon logic \cite{shulaker2017three,zidan2018future}. This approach show the possibility to have high package densities that could be used to scale up the neural network implementations. Another option is the use of semiconducting nanowires that demonstrate NVM properties\cite{sangwan2020neuromorphic}}.

Another post-CMOS strategy that is upcoming is the use of quantum computing. Recently, researchers started developing Noisy Intermediate Scale Quantum (NISQ) chips that aim to have 50-100 qubits \cite{ding2020quantum}. The possibility for quantum chips to run several operations in parallel, inspire the design for quantum neural network. Attempts are made to build  quantum neural network using silicon fabrication techniques, implemented on a NISQ chip with 170 control parameters in 2-by-5-millimeter area \cite{carolan2020variational}.

\section{Edge AI Chips}

{\label{352365}}

\subsection{Integrated Edge AI Platforms}
{The intelligence at edge has been a major driver in the recent development of AI chips. Table 1 shows a collection of most popular hardware solutions for
edge AI implementation available in the market today.} Most, if not all, of the commercial solutions of Edge AI hardware, are based on digital and mixed-signal logic and include concepts of
in-memory computing and parallel computations for speeding up the MAC operations. Like any hardware computing solutions, the AI chips differentiate in terms of performance issues resulting from  memory bandwidth, multiple I/O channels, memory architecture, modularity and scalability of the computing/embedded AI architecture, parallel and pipelined architectures, and ability to easily map AI architectures into AI chips.

\begin{table*}[!ht]
    \centering
       \caption{{Popular commercial Solution available for Edge AI hardware}}
 \scalebox{0.65}{   \begin{tabular}{p{3cm}|p{3cm}|p{4cm}|p{4cm}|p{4cm}|p{3cm}}
    \hline
    Solution & \multicolumn{4}{c}{Specification}\\ \cline{2-6}
             & Processor & Memory & Power & Speed & Size \\\hline
       Coral Dev Board   & TPU &  1GB RAM + 8BG flash memory & 0.5W per TOPS (2 TOPS per W), USB (5V) &  4 TOPS  & 48mm$\times$40mm$\times$5mm  \\\hline
       Jetson   TX2  & Embedded   GPU & 8 GB RAM + 32 GB Flash Memory  & 7.5-15W &  1.3 TFLOPS  & 87mm$\times$50 mm\\\hline
       Jetson   Nano & Embedded   GPU & 4 GB RAM + 16 GB Flash Memory & 5-10 W &  0.5 TFLOPs  & 69.6mm$\times$45mm\\\hline
       Intel Movidius Neural Compute Stick & High Performance VPU (vision processing unit) & 1 GB RAM + 4GB Flash memory &  1W & The USB3 interface- Super Speed (5 Gbps) or High Speed (480 Mbps) modes& 72.5mm$\times$27mm$\times$14 mm\\\hline
        Raspberry Pi 4& Quad core 64-bit ARM-Cortex A72   & 1, 2 and 4GB LPDDR4 RAM options & 3.4 watts& 1.5GHz & 85.6mm $\times$ 56.5mm\\\hline
        OpenMV Cam & low power camera board for visual data processing & 512KB RAM + 2 MB Flash Memory & Consumes 100 mA while idle and 140 mA when processing images, 3.3V & 480 MHz & 45mm$\times$36mm$\times$30mm \\\hline
        SparkFun  Edge  Development  Board  & 32-bit ARM Cortex-M4F processor & 384KB SRAM + 1MB Flash Memory &  6uA/MHz, 1.8V - 3.6V &  48MHz CPU clock, 96MHz with TurboSPOT & 40.6mm$\times$ 40.6mm$\times$ 8.9mm, \\\hline
        BeagleBone  AI & 2x ARM Cortex-A15&  1  GB  RAM  +  16  GB  Flash Memory&  Power up under 100mA (~500mW), USB type-C 5 volt, 2.5 amp  &1.5GHz & 45mm$\times$36mm$\times$30mm \\\hline
       Versal VC1902 ACAP Board &  Versal VC1902& DDR4 UDIMM 8GB, LPDDR4 Component 8GB & 12V Wall Adapter and ATX PCIe 8pin (6+2) Connector& 33 to 133 peak TOPS (for INT8 and INT16 operations), or 8 TFLOPS for FP32 operations & 190mm$\times$241mm \\ \hline
   %     Microsoft Brainwave & FPGA, 39.5 TFLOPs \\
   %    ARM ML & FPGA &  1 GB RAM & & 4 TOPs/W \\
   %     IBM's TrueNorth processor & SNN processor, 65mW\\ \hline
    \end{tabular}}
 
    \label{tab:my_label}
\end{table*}

%The MAC operation is required for most neural network implementations as they are required for the weight summation of inputs in each neural network layer.~ Such weighted summation operations are used for modeling neurons and for filters such in convolutional neural networks. The digital implementations also require extensive memory management and higher data transfers between the MAC and memory units.

%The growing interest in edge devices leads to the development of low-power neuromorphic chips for edge computing. Several neural systems for edge computing, such as Intel Movidius Neural Compute Stick, Jetson Nano~and Coral Dev Board, are already available on the market.

Current research in neuromorphic computing and neural chips focuses on
spiking neural networks and spike-based systems~\cite{Sung_2018},
probabilistic computing~\cite{Teh_1989,Myers_1988} and ReRAM based networks
(Figure~{\ref{654862}} (c)). Such systems involve the
development of analog and mixed-signal architectures which can also be
useful for near-sensor processing, rather than being purely digital based designs.
Spike-based systems rely on the spike-timing-dependent plasticity (STDP)
principle of the neural computations~\cite{Sj_str_m_2010}. The good
examples of SNN-based chips are TrueNorth by IBM and Loihi by Intel.
Probabilistic computing relies on the fundamentals of uncertainty and
noise in the natural data processing. ReRAM-based networks mostly
focus on the crossbar-based multiply and accumulate operations, while
accelerating the processing with analog domain computation using
low-power emerging devices. The advantages of such systems are small
on-chip area, high-density computations, and low power consumption. However, the ReRAM based neuromorphic systems suffer from endurance and robustness issues, limited precision, and device variability as the technology is relatively new. Also, the development and fabrication cost of such a system can be high, as the technology has not become mature yet.

The ability of the chips to handle large scale neural network architectures vary. As the number of neurons and layers in the architectures increases, it becomes challenging to run inference problems in real-time. Mainly, optimization of an architecture for a general-purpose AI hardware is not an easy problem, as the optimal configuration varies from one problem to another. Another well-known
problem is that with an increase in architecture size, there is an increased need to store the weights and perform MAC computations.
Almost, all general-purpose AI hardware is limited in both memory and computations, that limit and prevent implementation of large scale
neural networks.

\subsection{Types and purpose}

{\label{173872}}

The Edge AI and Edge computing are not the same. The purpose of edge computing systems is to off-load the computing from a cloud to devices
on edge. {In contrast, Edge AI is a subclass of edge computing that
focuses on providing AI functions to the edge devices.} The edge devices having AI functions can lead to faster real-time computations, saves
energy, consume lesser bandwidth, and can improve the security of intelligent data processing.
\begin{figure}[!ht]
\centering
\includegraphics[width=70mm]{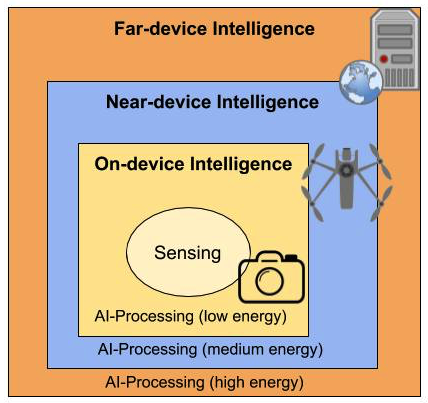}
\caption{{Edge AI implementation categories
{\label{485831}}%
}}

\end{figure}

There are different possibilities for have edge AI implemented as shown in Fig. {\ref{485831}}. This can be classified in hardware contexts as: (1) On-device intelligence, (2) Near-device intelligence, and (3) Far-device intelligence.

%\begin{enumerate}
%\tightlist
%\item
 % On-device intelligence
%\item
%  Near-device intelligence~
%\item
 % Far-device intelligence
%\end{enumerate}

\subsubsection{On-device intelligence}

{\label{702024}}

On-device intelligence can be defined as the set of methods that embed neural networks within the sensors and processing units. This can be a
single-chip or multi-chip solution. A good example of this is in-pixel processing or in general in-sensor processing, where the sensors contain neural circuits that enable them to perform intelligent operations. Another example is a smart camera, where the sensor and co-processor can be separate units, however, the camera is a single device that performs both sensing and processing.

\subsubsection{Near-device intelligence}

The near-device intelligence hardware performs intelligent computations
outside of the device, but they are physically near to the device. In
these devices, sensed data is stored and basic computations performed
with the device. However, the intelligent processing is done in another
device that is close to the sensing device. A common example is the
drones with camera, that collect the images and send them to a near-by
processing station. {The processing station performs intelligent
processing on images trying to identify objects, and perform real-time
segmentation and tracking.}

\subsubsection{Far-device intelligence}\label{far-device-intelligence}

The far-device intelligence hardware is a cloud or server-based systems
where intelligent data processing occurs away from the sensing device.
The device senses the real-time signals and then transmits it for
further processing in a remote server. In this system, some basic edge~
AI processing can be done, while the higher-level abstractions and
complex processing are performed in a high-performance computing
platform.

\section{Recipes to build AGI Chip
blocks}

{\label{738216}}

%\deleted{Autonomy and free will is a key aspect of building consciousness}
{Incorporating various priors (as defined in Section II) are important to build AGI
chips that resemble human intelligence.} There are many possible approaches to arrive at AGI chip
capability. At the highest level, each neural block of the chip should
be trainable and configurable in many different functions. {None of the AI chips today puts together the different cognitive priors required for mimicking human-like general intelligence. Following are
the main practical high level concepts that need to go in the design of AGI chips: (1) Ability to reconfigure the connectivity between neurons, neural network layers and that of neural system, (2) Randomness in the strength of connections and weight assignments to incorporate generality, (3) Presence of various types and forms of feedback and feedforward networks connections, and (4) different ways of learning the networks. Collectively, these high-level concepts should aim to {embed} the cognitive priors such as neuro-sensory, learning and knowledge as part of the built-in-test for the chip.  }

\subsection{Reconfigurability}

{\label{623112}}

The ability of the brain to reconfigure is based on several
factors~\cite{Betzel_2017,Puxeddu_2020}. Often memories and events are strongly
linked to the ability to forget, store, and filter information. The
reconfigurability is time-dependent and dynamic and {depends} on input
stimuli.

Consider the connectivity strength between any two
nodes~\(\{ n_1, n_2 \}\) as~\(c\) then, any input
stimuli~\(s_1\) can influence the change in the state of
connection from~\(c(t_1)\) to~\(c(t_2)\), however, not
necessarily changing the network decision~\(d_o\). Any change
in the network decision~\(d_o\) due to the change in~input
stimuli~\(s_1\) implies that the neural network learns a new
behavior.~ There can be a direct or indirect relationship between the
stimuli and the ability of the network to be reconfigured. The
flexibility of the network to be reconfigured offers the development of
multiple network structures from the same common blocks.

The memories and computations co-exist in neural
networks~\cite{Destexhe_2004,Machens_2012}. Unless the reconfiguration of the network
is possible in an easy manner, it will be difficult to encode the input
stimuli and capture the variations of the encoded stimuli within the
network. {A fully connected network would involve a connection between the neuron within the layer and between the layers. The connections will also be
fed back to the neuron within the same layer and will connect to the neurons in the previous
layers. The ability to connect, disconnect, and reconnect the
neurons can define the architecture, and the memory functions of the
network.}

\subsection{Randomness~}

{\label{404833}}

%The idea of randomness is closely linked to the information processing mechanisms in the human brain. It is understood that as the brain learns stimuli over a period of time, it shows stronger functional connections while having a decrease in randomness within the networks~\cite{Smit_2012}. 

%The randomness in the brain is also attributed to the ability of humans to be creative\cite{Sequeira_2001,STANISH_1986,Gorban_2017}.~

{It is understood that as the brain learns stimuli over a period of time, it shows stronger functional connections while having a decrease in randomness within the networks~\cite{Smit_2012}. In contrast, it has been seen in artificial neural networks that random dropouts can
benefit learning, and improve the robustness of the
networks~\cite{Poernomo_2018,Ko_2017,Wang_2019}. The amount of randomness can have a major impact on the learning ability of the networks. As the randomness
increases, the learning will take a longer time, while it also helps to
understand newer information. The number of weights in a network is
limited, having a percentage of weights exhibiting random variations
alters the connectivity and decision space of the neural network. It is
very well understood that a young brain develops fully connected
networks and gradually lose those connections as learning
occurs~\cite{Supekar_2009,Menon_2015,Mittner_2013}.~ As aging occurs to the brain, they become
exposed to various rewiring of the brain and become less
efficient~\cite{Achard_2007}. The randomness in the brain is also
attributed to the ability of humans to be creative\cite{Sequeira_2001,STANISH_1986,Gorban_2017}.~ It can be assumed that the ability to
randomly rewire the parts of the brain to process newer information
makes the information processing in the brain robust even with a reduced
efficiency as the aging occur.}~

The randomness also occurs in the input encoding of the stimuli, the
neuron thresholds, and the charge-conductance mechanisms. As each neuron
has minor variation in size, volume and structure, they respond
differently to the same input stimuli yet makes the overall architecture
robust for learning new information, providing evidence to the
importance of randomness in the intelligent design. Analogous to this,
the in-memory computing\cite{Mohamed_2020}, and analog
computing~\cite{2010} often naturally observe randomness in the
artificial neurons, drawing similarities to the biological neurons. The
dot-product computation implemented as multiply and accumulate operation
with crossbar architecture \cite{Vourkas_2014} is a good example of
this. The weighted summation of input voltages when reading as output
currents in the crossbar often has errors, resulting from the
conductance variations of the crossbar nodes or that from the leakage
currents and parasitic effects of the switches. While these errors exist, the
crossbar along with activation functions in a neural network
architecture compensates for these errors. The errors can be compensated
by tuning the conductance of the nodes, and robustness improved by
adjusting the thresholds and shape of activation functions.

\subsection{Role of feedback and feedforward
loops}

{\label{411552}}

{Both feedback and feedforward networks exist in the human
brain~\cite{Seidler_2004,Mejias_2016,Lillicrap_2019,Lillicrap_2020}. The studies in the past have argued for the
feedforward networks to play a major role in the prediction mechanisms,
while the feedback to be important to generalisation and learning abilities.}
Inferences and decisions often require information processing through
the layered neural networks in the shortest possible time. It should be
noted that all information-processing mechanisms in the human brain are
dynamic and real-time. Once an inference is made, some forms of actions
follow; needless to say even inaction is an action. The actions are also
influenced by various cognitive priors, and the interplay between the feedback and
feedforward networks are essential. As such most neural networks that
contribute to intelligence in the brain are recurrent, stochastic,
dynamic, and deep.

It can be also argued that the number of feedback and feedforward
networks in the human brain far exceeds than that of other animals. {As
the density of neurons increases, they lead to many different possible
combinations for feedbacks, which contributes to the higher holistic
sensing ability of the humans - be it vision, speech, smell, or touch.
Although many animals have one or other heightened forms of senses,
humans can collectively use different senses to build
awareness. The self-awareness originates from the basic responses of the
senses and the perception of the world that is mentally visualized.
Feedbacks in this sense, are essential for building newer learning priors and knowledge paths.} %\selectlanguage{english}
\begin{figure}[!ht]
\begin{center}
\includegraphics[width=1\columnwidth, trim={0 2.5cm 0 0},clip]{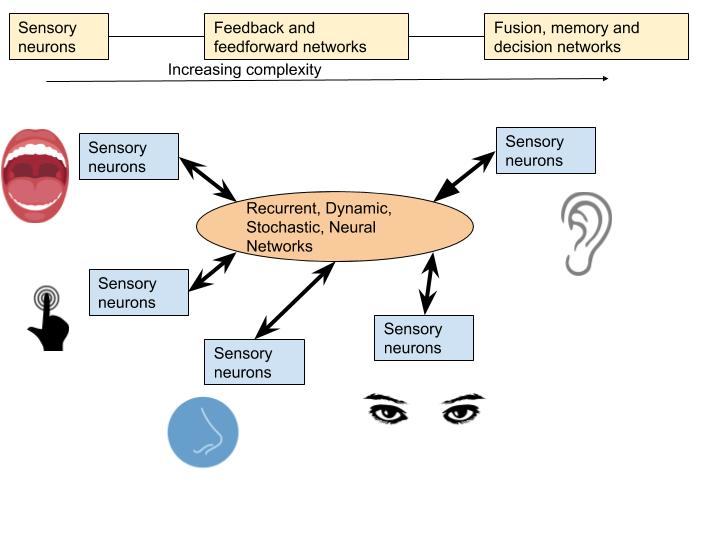}
\caption{{Increasing complexity with increasing sensory feedbacks and decisions
{\label{486770}}%
}}
\end{center}
\end{figure}

{Figure~{\ref{486770}} shows the overall high-level
structure of the real-time neural system that mimics general
intelligence. The AGI system should be able to integrate the motor-sensory priors that are basic
senses humans are used to and be able to sense and extract the features
through neural encoding mechanisms. The sensory neurons after processing
the stimuli send it to a series of neural networks that have feedback
and feedforward networks. The local decisions from information obtained
from each of the senses are further processed by the network of decision
networks that construct the memories and have the ability to fuse the
information to make meaningful conclusions. The neurons that uses such meta-learning priors are considered
as the building blocks, and as the information processing navigates
through to the decision networks, the mechanisms become complex giving
form and structure to the captured information. The robustness and generality of such a system is assessed using knowledge priors, that helps to qualitatively compare the performance of the AGI with that of human like intelligence.}

\subsection{Learning~}

{\label{530513}}

Learning is essential for human survival~\cite{Snigdha_2017,Peven_2019}. For surviving the challenges posed by the changing environment, setting a high level of motivation and
reward is important~\cite{O_Doherty_2006,Beninger_2018}. Rewards and goals drive the learning process
and often help to become efficient. In artificial neural networks, the
rewards are set as part of the objective function, with a reduction in
error as a way to make better decisions. The weights of the neural
network connections are updated or changed based on the final output
errors, which are propagated from one side of the network layer to
another. {The reduction in the errors can be achieved by updating and
readjusting the weights from one layer to another layer. This type of learning can be achieved in several different ways and algorithms, and cumulatively comes in the broader definition of meta-learning ability. }

The brains are unique~\cite{Miller_2012} and the learning process in the
human brain is still not fully understood. Every part of the brain has a
function, however, when a part of the brain becomes damaged, very often
some other part of the brain compensates for that loss. {This is possibly
due to the high level of interconnection between the neurons across the
brain, which gives many possibilities of translating biological meta-learning
into different forms of learning algorithm.}

It can be also noted that the nature of experiences and inputs that the
brain receives has a lot to do with the intelligence
abilities~\cite{Kolb_2013}. The highly plastic nature of the brain
allows for this, whereby, learning occurs in real-time and is always
online. The learning responds to previous memories, events, and updates
the inferences based on the newer experiences. Eventually, the outcomes
are tied to events making the learning algorithms change the
expectations on objectives over a period.

Not one learning algorithm would be sufficient. The skill attainment
would require the presence of mind that requires extensive use of
working memory. The way the updates are done in working memory is not
the same as that requiring creative forms of expressions and those involving executing the functions. It will be important to note in the
design of AGI systems that one universal algorithm for learning might
not be the solution, instead, the learning needs to be addressed at the
level of the neuron, neural networks, and network of systems in
different ways.~

\subsection{Scalability}

{\label{989418}}

The biological neurons are known to be on the scale of billions\cite{carter2019}. The neurons in the brain architecture
consist of complex connections between the neurons enhancing the functional purpose. The artificial neurons if it was to be useful would require an architecture that is scalable both structurally as well as
functionally.

{Almost} all the hardware devices and materials used for building neurons have electrical and cost limitations\cite{Kudithipudi,Wouters_2019}. The ability to replicate a module from one to another is an important aspect
of hardware design. Often a hierarchical approach to design helps to
reduce the complexity of troubleshooting and also helps to bring out the
designs faster.~ It can, therefore, be taken as an important principle in
the scalability design of the neural network. However, most neural networks are
fully connected within a layer and between the layers, which makes
signal propagation and electromagnetic issues important for neural
network designs.

While a weak AI chip is focused on solving problems with one type of
neural network, an AGI chip aims to incorporate generalisation that can emulate multiple neural
networks and functions. The aim of such AGI chips is to be fully reconfigurable such that various neural blocks can be configured to obtain specific neural
functions and learning approaches such as convolutional neural networks for object recognition
from images or long short term memories for processing and predicting
the weather using time series data. Collectively, they may be binded by  global learning mechanisms and fusion strategies. ~

\section{Architecture Blocks for AGI chip}

The functional properties outlined in Section \ref{738216} are required to put together the AGI chip design. Extreme parallelism, hierarchy, and modularity is a key aspect to speed up the design of the AI chips. Assuming these properties are taken care of from an algorithmic and hardware perspective, we list the main hardware blocks that form an AGI processor core. 

\begin{figure}[!ht]
    \centering
    \includegraphics[width=90mm]{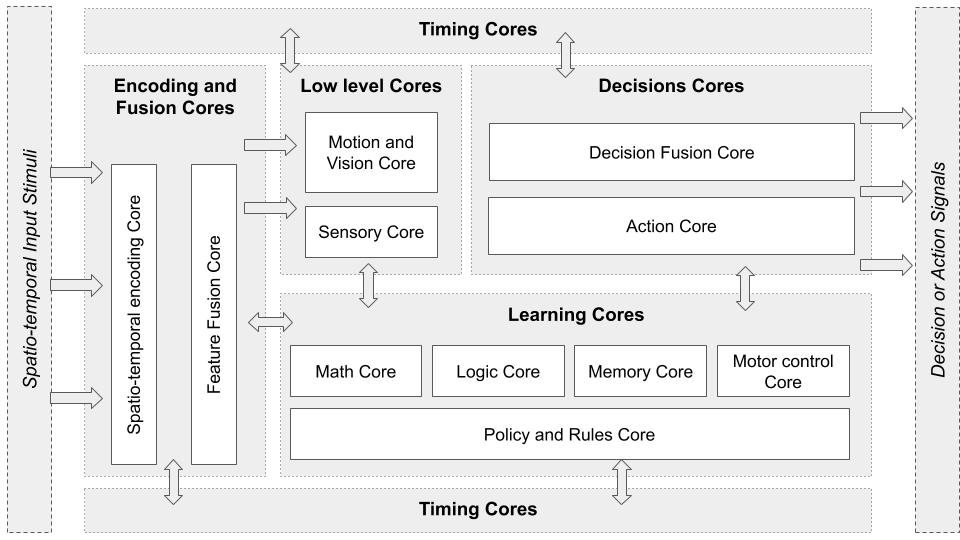}
    \caption{{The block diagram of a AGI processor core.}}
    \label{fig:agiblock}
\end{figure}

Various blocks of possible AGI chip is shown in Fig. \ref{fig:agiblock}. The system assumes that all signals from the sensors are time-dependent and have natural variations to it. \subsection{Encoding and fusion cores} The signals are in analog or digital form or in combinations of it. The Spatio-temporal block performs the probabilistic encoding of the input signals. The encoding can be implemented with a variety of techniques such as Spatial Pooler in HTM, LSTM, SNN encoding, etc. When passed through a set of encoders, the signal can be fused using a feature fusion core. The commonly used feature fusion approaches include wavelet-based fusion, concatenation, weighted addition, or the multiplication of features or using a neural network to combine the feature combination.

\subsection{Low-level cores} The sensory core and motion core forms the low-level cores required for the chip. The sensory core identifies the nature of the input signals as belonging to the five known human senses. {The sensory encoded signals from haptics sensors, olfactory sensors, gustatory sensors, speech and vision sensors, and vestibular sensors would be processed with this sensory core. These low level features could be processed in parallel to further encode and extract most relevant parts of the signals.} It also builds the attention features required to focus on objects or regions of signals needed for an application. For example, recognizing a human's activity in a scene requires the attention of that specific human and activities as identified through visual and audio cues. The motion core is necessary to build the low-level visual features, capture the depth of the scene, build the visual perception of surroundings, and capture the sensory signals' changes with respect to time.

\subsection{Learning cores} The logic core, math core, memory core, motor control core, and policy/rules core take part in the learning of features and fine-tuning the predictions.  The implementation of various learning algorithms, such as backpropagation, evolutionary computations, and optimisation requires modules for accelerating mathematical and logical computations.  In an online learning scenario, it is required to control the sensor feedback, such as in images to adjust the zoom, direction, and view angles, which is done using a motor control core. The intermediary variables, interrupt routines, and temporary storage is handled by the memory core, which can also be used as an in-memory computing unit. The memory core consists of both non-volatile memories for in-memory computations and volatile memories for dynamic storage. The policy and rules core keeps track of the conditions required for training the AGI system. Since an AGI system can have multiple learning algorithms and multiple neural network implementations running in parallel for different applications, it becomes essential to keep track of rules and policies specific to a given application.

{The learning core can be used for implementing higher level cognitive tasks such as for natural language processing, affective and psychological reasoning. The cognitive processing tasks are computationally expensive and uses the combinations of maths, logic,  memory and rules, which can be accelerated with AGI learning cores. Since many of the higher level cognitive tasks extensively uses large amounts of data for training the models, multiple AGI processor cores will be required. Furthermore, processors with multiple AGI cores will be a natural extension to implement ever increasing demands of higher level cognitive tasks.}

\subsection{Decisions cores}
Most real-time AGI applications acquire signals from multiple sensors, which require processing across numerous neural networks. The decisions from various sensory processing networks, such as visual cues, audio cues, etc, require consolidation through a decision fusion core, followed by taking appropriate actions for learning or inference. The action core will further process the decisions to send signals to indicate the execution of a set of actions required for the specific application. 
\subsection{Timing core}

The timing core organises the information flow in a sequenced order relating to events detected by the decision core.  The timing core is responsible for the AGI clock and is subjective to the actual time. In other words, the time is subjective to the succession of events that occur during a period of time for a given application. The sequencing process can be done by using custom made algorithms supported by learning cores and decision cores.

The AGI processor core from the heart of the hardware for AGI system helps accelerate the algorithmic response in real-time implementations. Multiple cognitive priors and related algorithms can be implemented with such cores. Like a general-purpose computing processor, the AGI core can be used in complex computer architecture with multiple AGI cores to achieve high parallelism levels. Several possibilities of higher-level architectures from a hardware standpoint is shown in \cite{9180545}, which can be one of the possible ways to implement with the proposed AGI processor cores.  

\section{AGI chip design - a lot of room at top and
bottom}

{\label{321616}}

\subsection{Funnel design flow}

{\label{723902}}

\begin{figure}[!ht]
\begin{center}
\includegraphics[width=70mm]{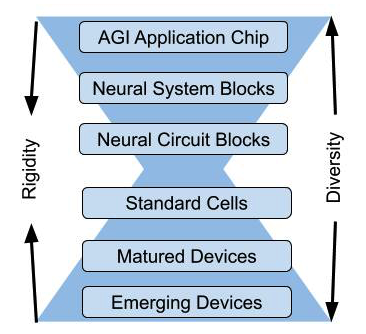}
\caption{{The funnel design flow for AGI Chip
{\label{993334}}%
}}
\end{center}
\end{figure}

As the push towards meeting Moore's law becomes challenging and costly,
the question of alternative devices and computing architecture
grows\cite{Shalf_2020,Schwierz_2020}.~ {It has been time and again quoted from the
device perspective that there is ``plenty of room at the bottom''\cite{feynman1960there} and
from the systems side ``plenty of room at the top''\cite{leiserson2020there}. This phenomenon is
reflected in the usage of ``more than Moore's law''. This leaves us with
a funnel design flow as shown in Fig~{\ref{993334}}.}

There are many new devices that over the years have shown promising results for designing and implementing AI chips. Some of the alternatives to CMOS based design include FDMOS, Carbon nano-tube
transitors, and FinFETs, and those that can reduce computations for
neural design include memristive crossbars with ReRAM, STTRAM, etc \cite{sangwan2020neuromorphic}.
Devices take many years of development to mature and cost a significant
amount of capital investments. Due to this reason, even if there are
many promising devices only very few make it through eventually to
industrial use. However, it should be noted that the search for the
perfect device that remains stable, reliable, which is easy to fabricate
and is compatible with popular integration processes continues as the
aims to pack more and more functionalities continue. This has also
opened up the field of quantum devices and computing as an integral part
of neural chip designs of the future.

The rigidity in the design and innovation is higher when designing
standard cells, both for digital as well as analog circuits. Any
performance errors in the standard cells will have a high impact on the
reliable functioning of the neural chips. It's at this stage that the
engineering fine-tuning and innovation steps in through a difficult
path. On one side, the circuit designers would need to be aware of the
device level issues, and on the other side, be able to build reliable
neural blocks that can withstand the test of time. Many standard cells
are used to build basic neural circuit blocks such as dot-product
computation, activation functions, and programming logic. The neural
circuit block also includes efficient designs of memory and state
transitions between the neural network layers. Several neural circuit
blocks are required to build neural system blocks such as convolutional
layers, hierarchical temporal memories, and long short term memories.
And a large combination of~neural system blocks is required for building
AGI chip targeted at general intelligence applications. The aim of such
AGI chips will be to ultimately match the wide spectrum of human
intelligence, leave plenty of room for innovation, development, and
applications.

\subsection{Automation of chip design}

{\label{834259}}

The design of the chips can be a time-consuming task. Automating the
design flow is an important task to speed up the design to the
manufacturing process, and for reducing the overall cost for developing
an AGI chip~\cite{Brandon}. Currently, in the digital chip market,
there are numerous tools for automating the ASIC design
flow~\cite{Uguen_2018} that can be used for developing an AGI chip.
Currently, in the digital chip market, there are a number of commercial
tools available for translating the high-level codes to digital logic.
The translations of mapping the logic to hardware come with different
efficiency and depend on the optimization methods used and technology
under consideration.~

Analog design automation is, however, much more challenging than digital
design automation. The device parasitics, process technology, signal
integrity issues, and thermal effects can significantly impact the
overall design of the analog circuit. It takes a high level of skill to
analyze through each of the issues and come up with a robust solution
for given process technology. The design that works for a particular
transistor node size does not more often work in a lower node size. This
makes the analog design challenging than digital design. One approach
used for automating the analog design for AI is evolutionary
optimization methods~\cite{Hakhamaneshi_2019,james2020}. This is an open area of
research and requires various levels of optimization to ensure the
selection of optimal hardware blocks for an AI processor design. For
example, in the case of emerging devices for analog AI chip design, the
variability in the devices will be much higher than traditional CMOS
based designs. This means the design automation for emerging~analog AI
chips \cite{james2020} such as using crossbar requires to consider
multiple objective functions to arrive at an optimal design. As the size
of the networks increases the search to find the optimal design becomes
challenging both at the schematic and physical design
levels.

\par\null

\section{How to test AGI Chips?~}

{\label{960251}}

An AI chip that implements AGI requires to be tested to ensure that
general intelligence functionality is achieved. This can be done by
testing the AGI chip in a real-time application involving conversations
that enable performing a Turning test.~ The ability of the AGI chips to
provide general intelligence in tasks that humans could do well is
indeed a well-known task. Likewise, it can be argued that the general intelligence tasks may or may not work successfully for humans is also
equally important.~

There are several aspects to test the effectiveness of the AGI chip. {The use of knowledge priors assist in testing and comparing the AGI chip functions with that of human intelligence.  Some of the functional test that addresses knowledge priors are outlined using algorithmic and philosophical approaches as shown in \cite{monett2020introduction, wang2012theoretical, hernandez2017measure}. Beyond functional tests, the AGI chips also would need to undergo electrical tests for ensuring hardware reliability.
This can be grouped into functional, and electrical tests, to test the AGI chip and the AGI system build with it. Obviously, such system can be a combination of software and hardware routines, with core aspects of AGI functions implemented on the hardware. The example
set of tests that broadly covers aspects of intelligence priors to chip performance are proposed below:} 

\subsection{Functional test 1: Dating test - conversation
game}

{\label{268495}}

The process of two humans engaging in a romantic date, and the ability
to have an engaging conversation is a mark of general intellectual
ability. AGI system should be able to consider effectively following factors:(1) make a wise selection of dating activities, (2) engage and communicate effectively, and (3) politeness and manners should have adhered.

AGI system will be presented as a real human using a chatbot or virtual
reality platform. The date will be introduced to the AGI bot as a real
human and asked to rate the AGI on three of the factors. As long as the
human is able to find the AGI bot personality attractive, and ask for
another date to meet in person, it will mark as a success. The test can
be repeated with several individuals to validate the strength of AGI. 

\subsection{Functional test 2: Sports commentary test - vision
game}

{\label{636353}}

General intelligence allows humans to follow rules and make practical
estimations based on a situation. This is very common during a sports
game. There are certain rules to follow, and at the same time,
improvisation is often required to win the game. When we watch a sports
game, we often make use of the majority of cognitive functions such as
hearing, language interpretations, decision rules, estimations, and
extensively use human vision. In this task, the AGI system should be able to: (1) recognize the sports through visual and auditory means, (2) be able to understand the rules and guidelines, and (3) engage the viewers through a wide set of conversations.

The viewers should not be able to recognize the difference between human
commentary and AGI commentary. The judgment of who passes the test
will be based on cross-validating the human judgment across several human
subjects.

\subsection{Functional test 3: Mentalist test - emotion
game}

{\label{704842}}

Humans are good at reading gestures and behavior of the other humans,
and use those cues to predict the thoughts. The mentalist theories
suggest that the ability to read another person's mind is a natural
process and can start very early in humans. Many subtle emotions are
invariably understood by the human mind and are often expressed through
emotional gestures. In this task, the AGI system should be able to: (1) understand gestures from visions, (2) be able to read emotional details through various senses, and (3) be able to guess the thoughts of a human being and be ethically sensitive about it.

The mentalist test could be the ultimate test for the AGI system. If the
AGI system could predict the thoughts of humans using the known senses,
it will be able to uncover and prove many theories of intelligence. Ultimately,
such a system will be considered superior to human general intelligence,
thereby resulting in singularity. At the same time, the AGI system
should act ethically and not reveal the personal details without breaching the privacy and permission of the individual. Such systems
could be possible with the help of additional sensing capabilities such
as reading the electromagnetic signals of the brain or that involving heat signature, which are not effectively possible with human senses.

\subsection{Electrical test 1: Architecture
robustness}

{\label{777301}}

The electrical noise, device parasitics, process variations, and signal integrity issues can affect the overall performance of the designed chip. Any such variations will result in signal loss, and the design methods used will have a major impact on the robustness of the chip.

The AGI architecture would need to be tested for the following: (1) stability against security attacks by adversarial manipulation of
  the signals within the chip, (2) sensitivity, specificity, and accuracy the architecture relative to
  the signal changes within the chip during the inference outcomes, and (3) robustness of architecture in application to multiple problems. 

The general intelligence nature of the AGI chip should not be affected by natural variations in input signals. While noise in the signal can be manipulated by an external force, the system is allowed to be tricked, at the same level of accuracy as that of the human brain.~

\subsection{Electrical test 2: Chip
reliability}

{\label{167811}}

The reliability of the chip requires extensive analysis and testing for functional accuracy under extreme situations in which the chip will be
used. As AGI chip will be required more often in real-time settings, it
will be important to have built-in systems for in-chip and in-system
monitoring, run-time checking, and built-in self-test.

The AGI chips would need to be reconfigured more often than a regular weak-AI chip. This means fault tolerance and error-correcting mechanisms
would need to be incorporated within the chip.~The AGI chip reliability
would need to be tested for the following: (1) thermal, electromagnetic and mechanical stress, and resulting
  benchmarks on the failure rate, (2) ability to compensate for the failures and have various built-in tests, and (3) the reliability of packaging and system integration for multi-chip
  communications.

The AGI chips would need to be integrated into real-time electronic devices under various applications such as in robotics, satellites, automotive industry, and mobiles. This would mean the packaging of the chips becomes important to~robotics, automotive industry, and mobiles, in improving the reliability when exposed to realistic and harsh environmental conditions. The practical applications warrant
that the chip would have a high endurance allowing it to be reused from one device to another.

\section{Discussion and future
directions}

{\label{781725}}

{There are several open problems in AGI chip design and implementations.
Some of those include (1) reliability and reconfigurability of the chip,
(2) architecture design and scalability, (3) compatibility with older
devices, (4) 3D integration challenges, (5) low power design challenges on multi-chip systems, (6) energy, area, and speed efficiency, (7)
biocompatibility and multisensor integration issues, (8) design and test automation, and (9) functional integration of multiple cognitive prior blocks required for AGI.}

%AI research has been focused on solving specific tasks that resulted in the automation of tasks and speeding up the quality and quantity of products developed. The innovations in science and engineering have been accelerated due to AI-driven techniques. 

%Many traditional fields that did not use computing and automation started to extensively use

AI-trusted computing such as in drug discovery, medicine, health
services, governments, and chemistry, opens up several opportunities for AGI chips. Law enforcement uses face
detection and recognition, biometric matching, and tracking extensively.
The extensive use of AI has led to the questions of ethics and politics of AI. Many corporate and governmental bodies started to used AI approaches to perform market analysis, monitor activities, understand
user behavior, and develop policies for better management.

AGI chips can be used to build ethically and socially aware AI, that can be integrated into human society without having apocalyptic fears about
AI. The AGI could eventually be part of the solutions for providing better services, and judgments such as in automation of services in
government or politics or courts.~ Another emerging possibility is to
integrate the AGI chip with that of the neural networks of the brain.
The bio-interface, human-AGI interactions, and cyborg applications can
lead to collective intelligence abilities, to enhance human intelligence
abilities.

\begin{figure}
    \centering
    \includegraphics[width=70mm]{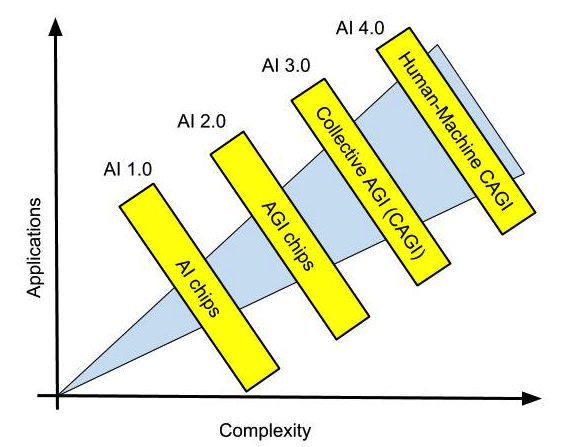}
    \caption{The future landscape in AI chip development}
    \label{fig:6}
\end{figure}

Figure \ref{fig:6} shows the four stages of AI chip development. Currently we are in AI 1.0 stage, with a major focus on specific applications. In the second stage of AI, the focus will be on building general intelligence chips, that could assist humans further in solving difficult social and ethical questions. The third stage could witness collective form of AGI, where the intelligence will expand in applications and complexity, currently not imagined by humans. The fourth stage and beyond, would see the integration of AGI chip with human brain, and ultimately allow humans to access all forms of electronic signals, and senses  available to machines. 

\section{Conclusion}

{\label{900499}}

This paper presented an overview of the emerging field of artificial
general intelligence from a hardware perspective. {The AGI systems, in
general, have been considered extremely difficult to build, and more
often the researchers gave up on the idea of building a realistic AGI.
The AI hardware support for building the AGI system has been limited and
has been proven to be a difficult problem even for a weak AI system. The development of algorithms and systems should be complemented with different types of AGI chips.
Emulating the brain networks through cognitive priors without a full understanding of the brain
mechanisms, also have been the major bottleneck in this area.}
Nonetheless, with the rapid progress in the AI-driven design
methodologies and the emergence of new devices offers possibility for building
full-fledged AGI chips. It is required to bridge
the knowledge gap between the brain sciences, psychology, computing,
materials psychology, computing, materials, and electrical engineering
among several other fields to address the questions of AGI systems.
Building AGI chips could be an efficient way to create energy-efficient
real-time neural computations and include accelerated collaborative
intelligence to compete with the general intelligence capabilities of
the human brain.~ While humans are limited by the number of sensory
mechanisms that support the general intelligence, the machines could
have a major advantage in using a wide range of sensing abilities such
as reading electromagnetic signals and visual spectrum that is not
accessible by the human brain, that can redefine the way we use and apply AI technologies.

%\section*{Acknowledgements}
%The funding support under industry research grant NeuroAGI-Biologics/2020/1 is acknowledged. 
%\FloatBarrier
%\balance
%\begin{spacing}{.9}
%\bibliographystyle{IEEEtran}
%\bibliography{bibliography/converted_to_latex.bib}
%\end{spacing}

%\begin{comment}
%\end{comment}


\end{document}